\pdfoutput=1 
\documentclass[letter, 10 pt, conference]{ieeeconf}  

\IEEEoverridecommandlockouts                              

\title{\LARGE \bf Towards Transferable Multi-modal Perception Representation Learning for Autonomy: NeRF-Supervised Masked AutoEncoder
}

\author{Xiaohao Xu$^{*}$
\thanks{X.H. Xu is with the Robotics Institute, University of Michigan, Ann Arbor, MI, USA. ({\tt\small xiaohaox@umich.edu}). This work was done when X.H. Xu was with the Research Center of Jueying Intelligence, Shanghai, China.
}
\thanks{The author would like to thank the suggestions from X.Z. Zhu, H. Tian, and L.W. Lu during the preliminary development stage of this work.}
}

\usepackage{times}
\usepackage{epsfig}

\usepackage{amssymb}
\usepackage{ifsym}

\usepackage{graphicx}
\usepackage{amsmath}
\usepackage[dvipsnames]{xcolor}
\usepackage[breaklinks,colorlinks]{hyperref}
\hypersetup{
    colorlinks=False,
    linkcolor=False,
    filecolor=False,      
    urlcolor=False,
    citecolor=False
}
\usepackage{subfigure}

\definecolor{lightgray}{rgb}{0.83, 0.83, 0.83}
\definecolor{forestgreen}{rgb}{0.13, 0.55, 0.13}

\usepackage{color}
\usepackage{latexsym}
\usepackage{CJKutf8}
\usepackage{soul}
 \usepackage{multirow}
 \usepackage{multicol}
 \usepackage{amssymb}
 \usepackage{algorithm} 
 \usepackage{algorithmic}
 \usepackage{booktabs}
  \usepackage{amsmath}
\usepackage{wrapfig}
  
\usepackage[dvipsnames]{xcolor}

\usepackage[utf8]{inputenc} 
\usepackage[T1]{fontenc}    
\usepackage{url}            
\usepackage{booktabs}       
\usepackage{amsfonts}       
\usepackage{nicefrac}       
\usepackage{microtype}      
\usepackage{xcolor}         

\usepackage{dsfont}
\usepackage{algorithm}
\usepackage{algorithmic}
\usepackage{subfigure}
\usepackage{graphicx}
\usepackage{epsfig}
\usepackage{amsmath}
\usepackage{wrapfig}
\usepackage{multirow}
\usepackage{amssymb}

\usepackage{comment}
\usepackage{color}
\usepackage{amsfonts}
\usepackage{url}
\usepackage{bm}
\usepackage{textcomp}

\definecolor{lightgray}{rgb}{0.83, 0.83, 0.83}
\definecolor{forestgreen}{rgb}{0.13, 0.55, 0.13}

\newcommand{\mypara}[1]{\vspace{1mm}\noindent\textbf{#1}}
\newcommand{\name}{NS-MAE}

\newcommand{\nus}{nuScenes}

\usepackage[capitalize]{cleveref}
\crefname{section}{Sec.}{Secs.}
\Crefname{section}{Section}{Sections}
\Crefname{table}{Table}{Tables}
\crefname{table}{Tab.}{Tabs.}

\definecolor{codegreen}{rgb}{0,0.6,0}
\definecolor{codegray}{rgb}{0.5,0.5,0.5}
\definecolor{codepurple}{rgb}{0.58,0,0.82}
\definecolor{backcolour}{rgb}{0.95,0.95,0.92}

\definecolor{MyGreen}{RGB}{0, 180, 0}
\definecolor{MyRed}{RGB}{180, 0, 0}
\definecolor{MyBlue}{RGB}{30, 0, 180}

\begin{document}

\maketitle
\thispagestyle{empty}
\pagestyle{empty}

\begin{abstract}
This work proposes a unified self-supervised pre-training framework for transferable multi-modal perception representation learning via masked multi-modal reconstruction in Neural Radiance Field (NeRF), \textit{namely} {\textbf{N}}eRF-{\textbf{S}}upervised {\textbf{M}}asked {{\textbf{A}}}uto{\textbf{E}}ncoder (\textbf{\name}). Specifically, conditioned on certain view directions and locations, multi-modal embeddings extracted from corrupted multi-modal input signals, \textit{i.e.}, Lidar point clouds and images, are rendered into projected multi-modal feature maps via neural rendering. Then, original multi-modal signals serve as reconstruction targets for the rendered multi-modal feature maps to enable self-supervised representation learning. Extensive experiments show that the representation learned via \name{} shows promising transferability for diverse multi-modal and single-modal (camera-only and Lidar-only) perception models on diverse 3D perception downstream tasks (3D object detection and BEV map segmentation) with diverse amounts of fine-tuning labeled data. Moreover, we empirically find that \name{} enjoys the synergy of both the mechanism of masked autoencoder and neural radiance field. We hope this study can inspire exploration of more general multi-modal representation learning for autonomous agents.
\end{abstract}

\begin{figure}[t!]
 \vspace{-4mm}
	\centering
	\includegraphics[width=0.48\textwidth]{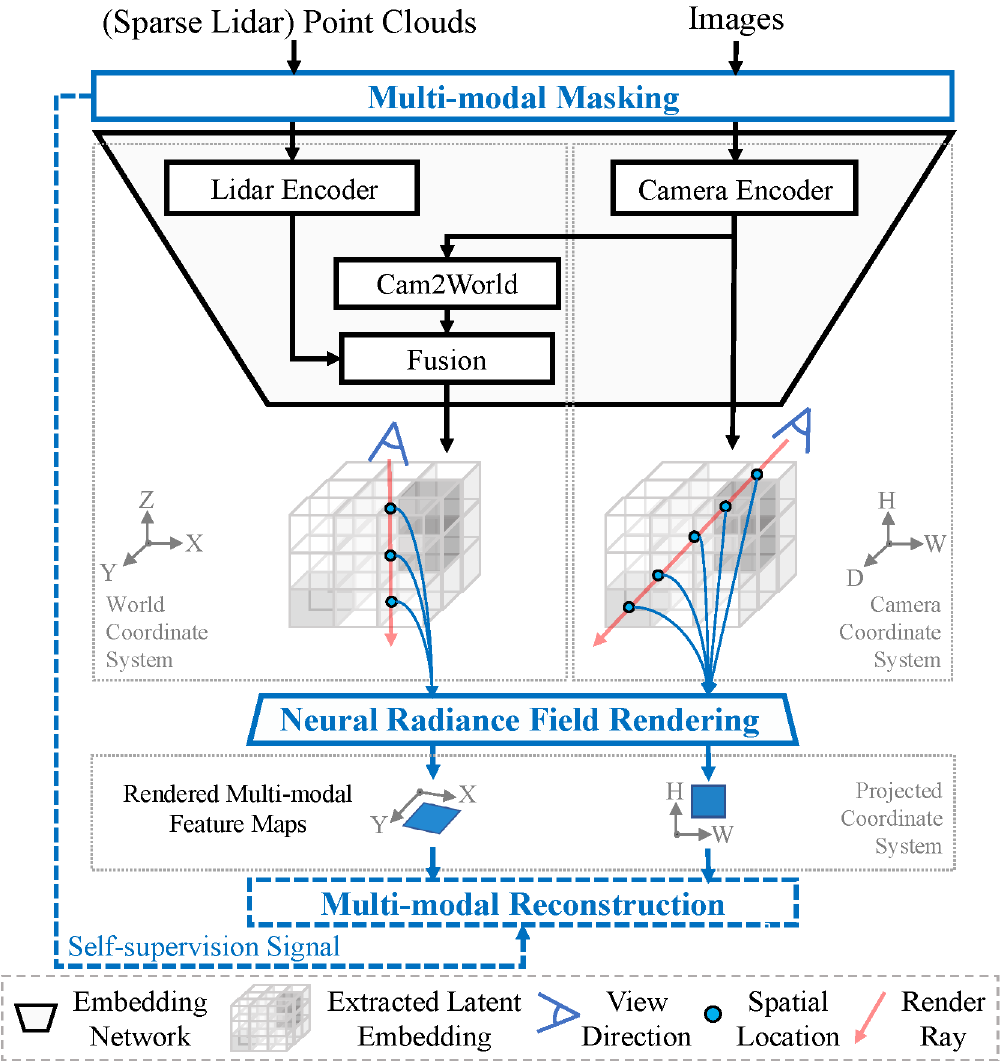} 
	\caption{\textbf{Overview of our unified pre-training framework for transferable multi-modal perception representation learning}. \textcolor{NavyBlue}{\textbf{Blue}} parts denote the proposed plug-and-play components for representation learning. Multi-modal inputs, \textit{i.e.}, sparse Lidar point clouds and images, are first partially masked out and sent to multi-modal encoders, \textit{i.e.}, the embedding network, for latent representation encoding. Then, given specific view directions and locations, the extracted embeddings are rendered into projected feature maps of various modalities, \textit{e.g.}, color and projected point cloud maps, via a differential neural volume rendering mechanism. Finally, the rendered feature maps are supervised by original inputs via self-supervised multi-modal reconstruction. }
	\label{fig:teaser}
\vspace{-4mm}
\end{figure}

\section{{Introduction}}

Toward robust autonomous driving, multi-modal perception~\cite{waymo,harley2023simple}, which aims to sense the surrounding scene by extracting and fusing representations from diverse modalities, is a crucial research direction. Meanwhile, with the booming development of modern network architectures~\cite{Vaswani2017AttentionIA,bommasani2021opportunities}, great efforts on transferable representation learning have been witnessed to relieve the appetite for data~\cite{he2022masked,bachmann2022multimae}. However, \textbf{pre-training for multi-modal perception is in its infancy.}

We first {review the widely-adopted {multi-stage fully-supervised training} paradigm for current advanced multi-modal perception models}~\cite{liang2022bevfusion,liu2022bevfusion,yang2022deepinteraction}. In this paradigm, the networks of Lidar and camera branches are first pre-trained separately in a fully-supervised learning way, and then the whole architecture is jointly fine-tuned. However, such a  paradigm is not scalable due to the {{scarcity of labels for supervision}}. Specifically, annotating paired images and extremely sparse Lidar point clouds for 3D perception can be cumbersome~\cite{FLAVA}, thus high-quality 3D labels are scarce. To relieve this problem, some self-supervised pre-training methods have been raised for single-modal perception models~\cite{dd3d,li2022simipu,CMKD,voxel-mae-wacv,voxel-mae} to enable label-efficient transfer. However, their optimization formulations are not unified and there is no work to explore transferable representation learning for advanced multi-modal perception models.

To this end, we go on to think about {\textbf{what should the ideal multi-modal perception pre-training framework for transferable representation learning be like?}} Inspired by successful pre-training frameworks for vision~\cite{he2022masked} and language~\cite{devlin2018bert}, apart from the effectiveness to boost performance, we argue that the following features are crucial:
\begin{itemize}
    \item \textbf{Unified}:  it should be generic to various settings, \textit{e.g.}, input modalities, architectures, and downstreams.
    \item \textbf{Scalable}: it should embrace the huge unlabeled yet valuable multi-modal data and potentially huge models.
    \item \textbf{Neat}:  it should possess a simple formulation to be adapted to more diverse modalities for perception.
\end{itemize}

With these high-level goals in mind, we decouple the design of the pre-training framework for multi-modal perception into two sub-problems: (1) \textbf{learning} transferable multi-modal representation; (2) \textbf{unifying} the optimization formulation for multi-modal representation.

    On one hand, to learn multi-modal representation in a self-supervised way, we get inspiration from the great success of Masked AutoEncoder~\cite{he2022masked,yang2023masked,bachmann2022multimae} (MAE) paradigm. Specifically, MAE~\cite{he2022masked} follows a mask-then-reconstruct paradigm and shows inspiring transferability on various downstream tasks. To take a step forward, we would like to explore the possibility of adapting MAE to learn transferable multi-modal perception representation, thus empowering advanced multi-modal perception models~\cite{liu2022bevfusion,liang2022bevfusion}.
     
     On the other hand, to unify the multi-modal representation optimization, Neural Radiance Field (NeRF)~\cite{aliev2020neural} provides a neat form to encode diverse physical properties, \textit{e.g.}, color and geometry, of the scene via differential neural volume rendering procedure. We consider NeRF as a useful building block for multi-modal perception pre-training for two reasons. Firstly, the imaging process of optical systems for perception, \textit{e.g.}, Lidar and camera, can be approximately modeled with rendering in the radiance field. Secondly, neural rendering unifies the multi-modal reconstruction via a neat and explainable physics formula. Thus, we introduce the rendering process of NeRF to perception pre-training, enabling unified multi-modal representation optimization.

Inspired by the remarkable attainments achieved by MAE and NeRF, we make a synergy of them and propose a unified self-supervised multi-modal perception pre-training framework (\name), which learns transferable multi-modal representations in a conceptually-neat formulation. As is shown in Fig.~\ref{fig:teaser}, we assemble the plug-and-play pre-training components (\textcolor{NavyBlue}{blue} parts of Fig.~\ref{fig:teaser}) to a typical embedding network of multi-modal perception models~\cite{liu2022bevfusion} (black parts of Fig.~\ref{fig:teaser}). During the pre-training, we first conduct modality-specific masking operations for multi-modal inputs, \textit{i.e.}, images and Lidar point clouds, separately. Then, we send the corrupted modalities to the multi-modal embedding network for embedding extraction. Specifically, we extract the embeddings that are generated and fused in two crucial coordination, \textit{i.e.}, the world and the camera coordination, for perception models~\cite{li2022bevformer,philion2020lift,saha2022translating}. Afterward, according to specific view directions and spatial sampling locations, the embeddings are further rendered into diverse projected modality feature maps via differential volume rendering. Finally, the rendered feature maps are supervised by original images and point clouds via reconstruction-based self-supervised optimization.

At last, we evaluate the transferability of multi-modal representation learned via \name{} for diverse multi-modal and single-modal perception models. 

In summary, this work has the following contributions:
\begin{itemize}
    \item We propose {\textbf{a novel unified self-supervised pre-training framework for multi-modal perception representation learning}}, \textit{i.e.}, \name, which is generic to both multi-modal and single-modal perception models.
    \item We enable both the \textbf{self-supervised learning} and \textbf{optimization unification} of transferable multi-modal representation via plug-and-play designs in the spirit of multi-modal reconstruction in neural radiance field.
    \item We employ \name{} to various advanced single-modal and multi-modal perception models and \textbf{verify the transferability of multi-modal representation derived via \name{}} on diverse 3D perception tasks with diverse amounts of fine-tuning data.
\end{itemize}

\section{Related Works}
\label{sec:related_work}

\mypara{Masked AutoEncoder} (MAE) paradigm~\cite{bao2021beit, chen2020generative,fang2022eva}, which originates from masked language modeling \cite{devlin2018bert} in natural language processing, learns transferable visual representation with masked signal reconstruction from partial observation. BEiT~\cite{bao2021beit} and its subsequent works~\cite{dong2021peco, peng2022beit, li2022mc,fang2022corrupted} leverage a pre-trained tokenizer and reconstruct masked image patches in the token space. MAE~\cite{he2022masked} and SimMIM~\cite{xie2022simmim} demonstrate that directly reconstructing masked patches in raw pixel space can also lead to good transferability and scalability. Other works perform reconstruction in a high-level feature space~\cite{zhou2021ibot,fang2022data,chen2022sdae} or handcrafted feature space~\cite{wei2022masked}. Enlightened by these methods, we would like to tap the potential of introducing MAE to multi-modal representation learning for perception in this work.  

\mypara{Neural Radiance Field} (NeRF)~\cite{mildenhall2020nerf} shows impressive view
synthesis results by using implicit functions to encode volumetric density and color observations. To improve the few-shot generalization ability of NeRF, data-driven priors recovered from general training data~\cite{yu2021pixelnerf}, other tasks ~\cite{jain2021putting,wei2021nerfingmvs}, or mixed priors~\cite{chen2021mvsnerf} are leveraged to fill in missing information of test scenes. For efficient NeRF training, some works~\cite{liu2020neural,neff2021donerf,wei2021nerfingmvs,deng2022depth} attempt to regulate the 3D geometry with depth. Considering the neat formulation of the neural rendering mechanism to optimize both the appearance and geometry, we introduce NeRF in perception pre-training.

\mypara{Perception Models for Autonomous Driving} include the following three main categories:
{(1) \textbf{\textit{Camera-only}}} models~\cite{lu2021geometry, Roddick2019OrthographicFT, liu2021autoshape, kumar2021groomed, zhang2021objects, zhou2021monocular, reading2021categorical, wang2021progressive,qin2019monogrnet, wang2021depth} initially focus on monocular 3D detection~\cite{kitti}. Thanks to the emergence of larger benchmarks ~\cite{Caesar2020nuScenesAM, waymo}, researchers start to study perception with range-view images~\cite{wang2022probabilistic, wang2021fcos3d, wang2022detr3d}. Later, LSS~\cite{Philion2020LiftSS}, which transforms perspective camera feature maps into 3D Ego-car coordinate, helps shift the perspective-view perception into bird's eye view (BEV) perception~\cite{huang2021bevdet, huang2022bevdet4d, reading2021categorical}, thus largely boosting the performance. (2) Early \textbf{\textit{Lidar-only}} methods either operate on raw Lidar point clouds~\cite{qi2017pointnet++,qi2018frustumPF,Shi2019PointRCNN3O,Yang20203DSSDP3,li2021lidar} or transform original point clouds into voxel~\cite{Zhou2018VoxelNetEL} or pillar representation~\cite{Lang2019PointPillarsFE, Wang2020PillarbasedOD, Yin2020Centerbased3O}.
Later, these two feature representations are unified in one single model{~\cite{Chen2017Multiview3O,Zhou2019EndtoEndMF,Shi2020PVRCNNPF}. 
(3) \textbf{\textit{Multi-modal}} perception models~\cite{Huang2020EPNetEP, sindagi2019mvxnet, Wang2021PointAugmentingCA,Yin2021MVP,Yoo20203DCVFGJ,bai2022transfusion,li2022deepfusion,yang2022deepinteraction}, which unleash the complementary power of multiple modalities, become the de-facto standard in 3D perception. Recent works \cite{liu2022bevfusion,liang2022bevfusion} propose more effective and robust multi-modal models via disentangled modality modeling.
Despite some attempts to explore pre-training for single-modal perception~\cite{PL,AMOD,DORN,dd3d,CMKD,li2022simipu,voxel-mae,liang2021exploring}, there is no trial for multi-modal perception models. Thus, we hope to explore representation learning for perception models with both single-modality and multi-modality in a unified form.

\section{Our Approach}
\label{sec:method}

\subsection{Overview}
\mypara{Problem Formulation of Perception Pre-training.} Given a tuple of images $\mathbf{I}=({I}_1, {I}_2, ...,{I}_N\in {\mathbb{R}}^{H\times W \times 3})$ collected from $N$ views with their corresponding camera parameters $(\mathbf{P}_1\mathbf{K}_1, \mathbf{P}_2\mathbf{K}_2, ..., \mathbf{P}_N\mathbf{K}_N)$ (where $\mathbf{P}$ and $\mathbf{K}$ denote camera pose and intrinsic matrix) and sparse Lidar point clouds $\mathcal{P}=\{{p}_i=[x_i,y_i,z_i,r_i]\in \mathbb{R}^4 \}_{i=1,...,t}$,\footnote{$x,y,z$ denotes the position of a point in the world coordinate space; the Lidar intensity $r$ is optional for the input but is used as common practice for the Lidar-based encoder of perception models~\cite{Zhou2018VoxelNetEL} to boost performance.} the goal is to design a proxy task to learn the parameter set, \textit{i.e.}, transferable representation, of an embedding network $\mathcal{\phi}_{emb}$, which can be used to initialize the parameter set of the downstream perception model $\mathcal{\phi}_{down}(\supset \mathcal{\phi}_{emb}$) for further fine-tuning. 

\mypara{Pipeline of \name{} Pre-training} (as is shown in Fig.~\ref{fig:teaser}) includes the following three key steps to enable the transferable multi-modal perception representation learning: 

(1) \textbf{{Masking}} (Sec.~\ref{Sec: Multi-modal Masking}): The inputs, \textit{i.e.}, images, and voxelized Lidar point clouds, are separately masked; 

(2) \textbf{{Rendering}} (Sec.~\ref{Sec: Radiance Field Rendering}): The embeddings encoded from masked images and point clouds are rendered into color and projected point cloud feature maps via neural rendering~\cite{aliev2020neural};

(3) \textbf{{Reconstruction}} (Sec.~\ref{Sec: Multi-modal Reconstruction}): The rendered results are supervised by the ground-truth images and point clouds via multi-modal reconstruction-based optimization.

\subsection{Training Architecture}
\label{Sec: Architecture Details}
As is shown in Fig.~\ref{fig:teaser}, the pre-training architecture, \textit{i.e.}, a typical embedding network of multi-modal perception models, comprises the following components:

\mypara{Camera Encoder} takes masked images as input and generates the perspective-view (PER) image embedding $\mathbf{e}_{I}^{PER}\in {\mathbb{R}}^{H/{\kappa \times W/{\kappa} \times D \times C}}$ ($\kappa$ denotes the down-sampling ratio). It can be implemented with Transformer-based~\cite{Vaswani2017AttentionIA,liu2021swin} or convolution-neural-network-based~\cite{he2016deep} architectures.

\mypara{Lidar Encoder} takes masked Lidar voxels as input and generates BEV embedding of Lidar modality $\mathbf{e}_{L}^{BEV} \in {\mathbb{R}}^{X \times Y \times Z \times C_L}$.  Following common practice of 3D perception~\cite{liu2022bevfusion,liang2022bevfusion}, it is typically implemented with VoxelNet~\cite{Zhou2018VoxelNetEL}.

\mypara{Cam2World} module is used to transform the perspective-view image embedding $\mathbf{e}_{I}^{PER}$ of camera coordination to BEV embedding $\mathbf{e}_{I}^{BEV} \in {\mathbb{R}}^{X \times Y \times Z \times C_I}$ of world coordination. Thus, image and Lidar embeddings can be aligned. We follow the implementation of lift-splat-shoot~\cite{Philion2020LiftSS}.

\mypara{Fusion} block fuses the BEV embedding of camera branch $\mathbf{e}_{I}^{BEV}$ and Lidar branch $\mathbf{e}_{L}^{BEV}$, to generate fused multi-modal BEV embedding $[\mathbf{e}_{I}^{BEV};\mathbf{e}_{L}^{BEV}]\in {\mathbb{R}}^{X \times Y \times Z \times(C_I+C_L)}$ via a simple concatenation ($[\cdot;\cdot]$).

\subsection{Multi-modal Masking}
\label{Sec: Multi-modal Masking}
\mypara{Image Masking.}  Following the MAE schema~\cite{he2022masked,xie2022simmim}, the original unmasked image $I$ is first divided into regular non-overlapping image patches. Then, a random binary mask $M\in {\{0,1\}}^{H\times W}$ is applied to mask out a large portion of image patches by replacing them with a learnable [MASK] token ($\in \mathbb{R}^{s\times s\times 3}$, where $s\times s$ denotes the patch size). Afterward, the image that is partially masked out is sent to the camera encoder for embedding extraction.

\mypara{Lidar Masking.} Inspired by the great success of MAE-style visual pre-training, some recent works~\cite{voxel-mae,voxel-mae-wacv,pang2022masked} extend it for point cloud pertaining. Similarly,  after transforming the input Lidar point cloud into its voxelized form, we mask a large fraction of non-empty voxels (70\% to 90\%). Then, the partially-masked voxels are processed in the Lidar encoder to generate the Lidar embedding.

\subsection{Neural Radiance Field Rendering}
\label{Sec: Radiance Field Rendering}
\subsubsection{Rendering Network}
\mypara{Vanilla Rendering Network} of NeRF~\cite{aliev2020neural} takes a set of posed images and encodes the scene with volume density and emitted radiance for the purpose of view synthesis. In NeRF, a rendering network $f$ maps a given 3D point $\textbf{x} \in \mathbb{R}^3$ ($\mathbb{R}^3$ denotes the scene’s world space) and a particular viewing direction ${\omega} \in \mathbb{S}^2$ ($\mathbb{S}^2$ denotes the sphere of directions) to the differential sigma field density $\sigma\in \mathbb{R}$ and RGB color $\textbf{c}\in \mathbb{R}^3$, like so: $f(\textbf{x}, \omega) = (\sigma, \textbf{c})$. 

\mypara{Conditional Rendering Network for Pre-training.} As our goal (representation learning) is different from the goal of vanilla NeRF (view synthesis), we leverage a different rendering network formulation. In specific, we additionally introduce a latent multi-modal embedding $\mathbf{e}$ from the embedding network of the perception model to the inputs of the rendering network $f$, like so: $f(\textbf{x}, \omega, \mathbf{e}) = (\sigma, \textbf{c})$. Thus, the gradient from differential rendering and the following reconstruction-based self-supervision stages can be back-propagated to the embedding network for end-to-end representation learning.

\subsubsection{Rendering Target}
\mypara{Vanilla Color Rendering}.
Given the pose $\mathbf{P}$ and intrinsic $\mathbf{K}$ of a \textit{virtual camera}, we shoot rays $\mathbf{r}(t) = \mathbf{o} + t\mathbf{d}$ originating from the $\mathbf{P}$’s center of projection $\mathbf{o}$ in direction $\omega$ derived from its intrinsic $\mathbf{K}$ to render the RGB color $\hat{\mathbf{C}}(\mathbf{r})$ via standard volume rendering \cite{kajiya1984ray}, which is formulated as:
\begin{equation} 
 \hat{\mathbf{C}}(t) = \int_{0}^{\infty} T(t)\sigma(t)\mathbf{c}(t) dt,
\label{equ:render_color_c}
\end{equation}
where $\mathbf{c}(t)$ and $\sigma(t)$ are the differential color radiance and density, and $T(t) = \exp(-\int_{0}^t \sigma(s) ds)$ checks for occlusions by integrating the differential density between $0$ to $t$. Specifically, the discrete form can be approximated as:
\begin{equation}
    \hat{\mathbf{C}}(\mathbf{r}) = \sum_{i=1}^{N}T_{i}(1 - \mathrm{exp}(-\sigma_{i}\delta_{i}))\mathbf{c}_{i}, 
\label{equ:render_color_d}
\end{equation}
where $N$ is the number of sampled points along the ray, $\delta_{i} = t_{i+1}-t_{i}$ is the distance between two adjacent ray samples and the accumulated transmittance $T_{i}$ is $\mathrm{exp}(-\sum_{j=1}^{i-1}\sigma_{j}\delta_{j})$.

\mypara{Multi-modal Rendering for Pre-training}. For multi-modal pre-training, the rendering targets can be extended to unleash the power of multi-modal data. In specific, apart from the color that reflects the semantics of the scene, the Lidar ray that captures 3D geometry in the form of point clouds is also a kind of radiance. 
Going beyond the differential RGB color radiance $\mathbf{c}(t)$ for color rendering in Eq.~(\ref{equ:render_color_c}), we introduce the differential radiance of \textit{any-modality} $\mathbf{a}(t)$ for multi-modal rendering. Specifically, the rendering of the projected feature map of \textit{{a}ny-modality} $\hat{\mathbf{A}}(t)$ is formulated as:
 \begin{equation} 
 \hat{\mathbf{A}}(t) = \int_{0}^{\infty} T(t)\sigma(t)\mathbf{a}(t) dt,
\label{equ:depth_render_m}
\end{equation}

\noindent Typically, to render the projected 3D point cloud feature map, \textit{i.e.}, 2D depth $\hat{\mathbf{D}}(\mathbf{r})$, the differential radiance of \textit{{a}ny-modality} $\mathbf{a}(t)$ can be set as the integration of the distance distribution field $\int_{0}^t dt$. Then, the discrete form to render the feature map of the projected 3D point cloud can be expressed as:
\begin{equation}
    \hat{\mathbf{D}}(\mathbf{r}) = \sum_{i=1}^{N}(T_{i}(1 - \mathrm{exp}(-\sigma_{i}\delta_{i}))\sum_{j=1}^{i-1}\delta_j), 
\label{equ:depth_render_d}
\end{equation}

\subsection{Multi-modal Reconstruction}
\label{Sec: Multi-modal Reconstruction}
\mypara{Vanilla Color Reconstruction Objective.} Given a set of rendering rays $\mathcal{S}_r$ passing through the pixels of the original image, the goal is to minimize the square of the $L_2$-norm of the difference between the ground truth color $\mathbf{C}(r)$ and the rendered color $\hat{\mathbf{C}}(r)$ of ray $r$:
\begin{equation}
    \mathcal{L}_{\mathrm{C}} = \frac{1}{|\mathcal{S}_r|}\sum_{r \in \mathcal{S}_r}\| \hat{\mathbf{C}}(\mathbf{r}) - \mathbf{C}(\mathbf{r}) \|_{2}^2,
    \label{equ:loss_c}
\end{equation}

\mypara{Multi-modal Reconstruction Objective for Pre-training.} Given a set of rendering rays $\mathcal{S}_r$ passing through the ground-truth target, \textit{i.e.}, \textit{pixel}, of \textit{any-modality}, we minimize the $L_p$-norm of the difference between the ground-truth view-specific projected feature map of \textit{{a}ny-modality} $\mathbf{A}(r)$ and the rendered result $\hat{\mathbf{A}}(r)$ of ray $r$ to the $p$-th power\footnote{$p$ is usually chosen as $p \geq 1$ to ensure convexity of the function.}:
\begin{equation}
    \mathcal{L}_{\mathrm{A}} = \frac{1}{|\mathcal{S}_r|}\sum_{r \in \mathcal{S}_r}\| \hat{\mathbf{A}}(\mathbf{r}) - \mathbf{A}(\mathbf{r}) \|_{p}^p,
    \label{equ:loss_mm}
\end{equation}
\noindent The \textbf{overall objective function}  jointly optimize the reconstruction for multiple ($K$) view-specific modalities:
\begin{equation}
\mathcal{L} =  \sum_{k=1}^{K} (\lambda_{k} \cdot \mathcal{L}_{\mathrm{A}_k} ) \label{equ:loss_final}
\end{equation}
 $\lambda_{k}$ indicates the coefficient to modulate the $k$-th sub-loss.

\renewcommand{\arraystretch}{0.5}
\begin{table*}[t!]
\small
\centering
\caption{\textbf{3D object detection results for Multi-modal Perception} model (BEVFusion~\cite{liu2022bevfusion}) on \nus{} \cite{Caesar2020nuScenesAM} $val$. The notion of modality: Camera (C), Lidar (L). \#Sweep denotes the number of Lidar sweeps. \#ImgSize denotes the resolution of images. The notion of class: Construction vehicle (C.V.), Trailer (Trail.), Barrier (Barr.), Motorcycle (Moto.), Pedestrian (Ped.), Traffic cone (T.C.). 
 	}
\resizebox{1.0\textwidth}{!}
	{
        \setlength{\tabcolsep}{6pt}
	\centering
	\begin{tabular}{l|ccc|cccccccccc|cc}
          \toprule
         \multirow{2}{*}{Method} & \multirow{2}{*}{Modality} & \multirow{2}{*}{\#Sweep} & \multirow{2}{*}{\#ImgSize}  & \multicolumn{10}{c|}{Per-class mAP} & \multirow{2}{*}{{\small{mAP}}} & \multirow{2}{*}{{\small{NDS}}} \\ \cmidrule(lr){5-14}
          &  &  &  & \small{Car} & \small{Truck} & \small{C.V.} & \small{Bus} & \small{Trail.} & \small{Barr.} & \small{Moto.} & \small{Bike} & \small{Ped.} & \small{T.C.}  &  & \\
         \midrule
          BEVFusion~\cite{liu2022bevfusion}& {LC} &{1} &{$128\times 352$}   & 81.1 & {37.4} & 12.3 & 59.0 & 31.5 & 64.1 & {46.7} & {28.9} & {80.3} & {63.1}  & {50.5} &{53.3}\\ 
        + \name & {LC} &{1} &{$128\times 352$}  & {81.6} & {40.1} & {13.9}& {59.8} & {30.1} & {64.7} & {{48.9}} & {{30.3}} & {{81.0}} & {64.4}    & \textbf{{{51.5}}} &\textbf{{54.7}}       
         \\         
         \midrule 
       \textcolor{black}{BEVFusion}~\cite{liu2022bevfusion} & {LC} &{9} &{$256\times 704$}  & 87.4 & {40.4} & {{25.7}} & 67.0 & {38.8} & 71.6 & 68.2 & {48.6} & {{85.5}} & {{74.5}} &  {60.8} &{64.1} \\
          + {\name} & {LC} &{9} &{$256\times 704$}    & {{88.1}} & {{45.9}} & {25.1}& {{68.8}} & {{37.2}} & {73.8} & {70.8} & {56.6} & {86.9} & {77.4} & \textbf{{63.0}} &\textbf{{65.5}}  \\ 

        \bottomrule
	\end{tabular}
       }
	
\vspace{-4mm}
	\label{tab:nuscene_det}
\end{table*}

\subsection{Training Setup and Details}
\label{Sec: Pre-training Setup}
 \mypara{Training Strategy.}  The embedding network (Sec.~\ref{Sec: Architecture Details}) is pre-trained for 50 epochs. We use AdamW~\cite{adamw} optimizer with a learning rate of 1e-4 and a weight decay of $0.01$. Following mainstream perception models~\cite{Zhou2018VoxelNetEL,liu2022bevfusion,liang2022bevfusion}, one-cycle scheduler~\cite{one-cycle} is adopted. Early-stopping strategy~\cite{prechelt2012early} is used to avoid over-fitting. The network is trained on 8 NVIDIA V100 with a total batch size of 16.

 \mypara{Masking.} (1) For image: the masking patch size is set as $4\times4$ and $8\times8$ for images with resolutions of $128\times352$ and $256\times704$; the masking ratio is set as 50\%. (2) For Lidar: the point cloud range is set as -54$\sim$54(m), -54$\sim$54(m), and -5$\sim$3(m) for $X, Y$, and $Z$ axes, respectively, and the voxel size is set as [0.075, 0.075, 0.2](m); the masking ratio of non-empty Lidar voxel is set as 90\% or in a range-aware manner~\cite{voxel-mae} for single-sweep and multi-sweep point clouds. 

  \mypara{Rendering.}  (1) For rendering view directions, we select two typical and critical views for 3D perception models, \textit{i.e.}, bird's eye view (BEV) ${\omega}^{BEV}$ and perspective view (PER) ${\omega}^{PER}$. (2) For the rendering network, we implement it with \textit{conv} layers. Specifically, the rendering network first transforms the embedding $\mathbf{e} \in \mathbb{R}^{D_1\times D_2\times D_3 \times D_C}$ \footnote{In the special case, when $\omega={\omega}^{PER}$, dimensions $D_1,D_2,D_3$ refer to $H,W,D$ axes of the camera coordination; when $\omega={\omega}^{BEV}$, $D_1,D_2,D_3$ refer to $X,Y,Z$ axes of 
the world coordination.}extracted from the embedding network into sigma-field feature volume ${V}_{{\sigma}}\in \mathbb{R}^{D_1\times D_2\times {D^{'}_3} \times 1}$ and color feature volume ${V}_{\mathbf{c}}\in \mathbb{R}^{D_1\times D_2\times {D^{'}_3} \times 3}$, which are then processed via rendering to derive the projected feature maps for further reconstruction-based optimization. (3) For discretized rendering functions, the parameter $\delta$ is approximately set as $0.2$ and $0.8$ for the rendering in BEV and perspective view. 

  \mypara{Reconstruction.} (1) For reconstruction targets, we leverage color field map $\mathbf{C}$ which corresponds to multi-view camera-collected images, perspective-view depth $\mathbf{D}^{PER}$ which is generated by projecting Lidar point clouds on perspective-view image planes, and BEV depth $\mathbf{D}^{BEV}$ which is generated by projecting the voxelized Lidar point cloud on the BEV plane. (2) For the set of rendering rays $\mathcal{S}_r$, it is constructed with rays emitted orthogonally to the BEV plane to render depth in BEV ($\mathbf{D}^{BEV}$), and it is constructed with rays passing through image plane to render color and depth in perspective view ($\mathbf{C}$, $\mathbf{D}^{PER}$). (3) For the parameter $p$ in Eq.~(\ref{equ:loss_mm}), we set it as 2 and 1 for color and depth, respectively. (4) For coefficients of the color, perspective-view depth, and BEV depth, we set them as 1e4, 1e-2, and 1e-2, respectively, to normalize the numerical value of diverse modalities.

\mypara{Implementation.} We implement the network in PyTorch using the open-sourced MMDetection3D \cite{mmdet3d}. Data augmentations mainly follow official Lidar and image augmentations for 3D perception models~\cite{liu2022bevfusion,liang2022bevfusion} except the ones that require ground-truth labels, \textit{e.g.}, database-sampler~\cite{Zhu2019ClassbalancedGA}. 

\begin{table}[t]
  \small
    \caption{\textbf{3D object detection results of Multi-modal Perception} models (VFF~\cite{li2022voxel} with various 3d detection heads, including SECOND~\cite{yan2018second}, PVRCNN~\cite{shi2020pv}, and VoxelRCNN~\cite{voxelrcnn}) on KITTI-3D~\cite{kitti} $val$. Results are for the car category and reported in $AP_{R40}$@0.7,0.7,0.7. $DB$ indicates database-sampler~\cite{Zhu2019ClassbalancedGA} data augmentation is used.  }
  \centering
  \setlength{\tabcolsep}{2.4pt}
  \resizebox{0.48\textwidth}{!}{
  \begin{tabular}{l|c|ccc|ccc}
    \toprule
    \multirow{2}{*}{Method} &\multirow{2}{*}{$DB$} & \multicolumn{3}{c|}{AP${}_{BEV}$} & \multicolumn{3}{c}{AP${}_{3D}$}\\ \cmidrule(lr){3-5} \cmidrule(lr){6-8}  
    {} && Easy & Moderate & Hard & Easy & Moderate & Hard \\
    \midrule
    VFF-SECOND & &  {91.71} &{85.77} & {83.54} & {87.25} & {76.44} & {74.02} \\
        + \name & &  \textbf{92.65} & \textbf{88.24} & \textbf{85.83} & \textbf{88.25} & \textbf{78.40} & \textbf{74.37} \\
    \midrule
     VFF-SECOND &$\checkmark$ &  {92.83} &{88.92} & {88.22} & {89.45} & {82.32} & {79.39} \\   
    + \name  &$\checkmark$&  \textbf{93.04} & \textbf{90.43} & \textbf{88.46} & \textbf{91.48} & \textbf{82.58} & \textbf{79.77} \\  
  \midrule 
     VFF-PVRCNN   &$\checkmark$&  {92.65} &{90.86} & {88.55} & {91.75} & {85.09} & {82.68} \\  
      + \name &$\checkmark$ &  \textbf{92.94} & \textbf{91.00} & \textbf{90.49} & \textbf{92.03} & \textbf{85.31} & \textbf{83.05} \\   \midrule
     VFF-VoxelRCNN   &$\checkmark$&  \textbf{95.67} &{91.56} & {89.20} & {92.46} & {85.25} & {82.93} \\  
      + \name &$\checkmark$ &  {95.57} & \textbf{91.69} & \textbf{89.23} & \textbf{92.51} & \textbf{85.59} & \textbf{82.95} \\   
    \bottomrule
    \end{tabular}
    }
    
    \label{tab:kitti}
    \vspace{-2mm}
\end{table}

\renewcommand{\arraystretch}{0.5}
\begin{table*}[t!]
\small
	\caption{\textbf{3D object detection results for Camera-only (top) and Lidar-only (down) Perception} models on \nus{} \cite{Caesar2020nuScenesAM} $val$. ${2\times}$ denotes doubled training epochs. \#Sweep denotes the number of Lidar sweeps/scans. \#ImgSize denotes the spatial resolution of images. }\vspace{-2mm}
\resizebox{\textwidth}{!}
	{
        \setlength{\tabcolsep}{6pt}
	\centering
	\begin{tabular}{l|ccc|cccccccccc|cc}
          \toprule
         \multirow{2}{*}{Method} & \multirow{2}{*}{Modality} & \multirow{2}{*}{\#Sweep} & \multirow{2}{*}{\#ImgSize}  & \multicolumn{10}{c|}{Per-class mAP} & \multirow{2}{*}{{\small{mAP}}} & \multirow{2}{*}{{\small{NDS}}} \\ \cmidrule(lr){5-14}
          &  &  &  & \small{Car} & \small{Truck} & \small{C.V.} & \small{Bus} & \small{Trail.} & \small{Barr.} & \small{Moto.} & \small{Bike} & \small{Ped.} & \small{T.C.}  &  & \\
         \midrule
          BEVDet~\cite{huang2021bevdet}& C &/ & $256\times 704$ & 35.3 & {15.7} & {2.7} & {18.2} & 5.5 & 31.0 & 19.8 & 18.3 & 26.9 & 46.0  & {21.9} &{29.4}\\ 
          + \textcolor{black}{\name}  &C &/ & $256\times 704$     &{{37.1}} &{{18.0}} &{3.6} &{{18.0}} &{7.0}  &{{41.3}}  &{{20.3}} &{{19.8}} &{28.1} &{{47.6}}         &\textbf{{24.1}} &\textbf{{32.1}}       
         \\  \midrule 
            BEVDet${}^{2\times}$~\cite{huang2021bevdet}& C &/ & $256\times 704$ & 37.1 & {16.4} & {2.7} & {18.9} & 5.8 & 42.0 & 20.7 & 18.3 & 28.3 & 49.0 & {23.9} &{31.8}\\ 
          + \textcolor{black}{\name}  &C &/ & $256\times 704$   &{{38.3}} &{{17.9}} &{4.5} &{{18.3}} &{8.1}  &{{47.1}}  &{{21.5}} &{{18.8}} &{29.4} &{{49.8}}   &\textbf{{25.4}} &\textbf{{33.9}}       
         \\ \midrule
 
          CenterPoint~\cite{Yin2020Centerbased3O}& L &9 & /    & 80.9 & 52.4 & 14.4 & {64.0} & 29.6 & 58.7 & 59.4 & 45.6 & 80.4 & 60.8  & {54.6} &{61.3}\\ 
          + {VoxelMAE}~\cite{voxel-mae}  &L &9 & /      &{{80.6}} &{{53.7}} &{13.7} &{{63.2}} &{29.2}  &{{61.1}}  &{{60.5}} &{{45.4}} &{80.4} &{{61.1}}  &{{54.9}} &{{61.4}}\\
          + {\name}  &L &9 & /      &{{81.2}} &{{53.0}} &{{14.7} }&{{63.7}} &{{30.2}} &{{60.0}}  &{{60.1}} &{{47.1}} &{{81.6}} &{{61.3}}      &\textbf{{55.3}} &\textbf{{62.1}}    \\    
        \bottomrule
	\end{tabular}
       }
        \centering

\vspace{-4mm}
	\label{tab:nuscene_det_merged}
\end{table*}
\begin{table}[t!]
        \caption{\textbf{BEV-Map segmentation results for Multi-modal and Camera-only Perception} models on \nus{} \cite{Caesar2020nuScenesAM} $val$. The notion of class: Drivable (Dri.), Pedestrian Crossing (P.C.), Walkaway (Walk.), Stop Line (S.L.), Carpark (Car.), Divider (Div.). Images of size $256\times704$ and multi-sweep (9) Lidar points are used as input.}
\vspace{-2mm}
\small\centering
\setlength{\tabcolsep}{0.7mm}
\label{tab:results:segmentation}
\resizebox{0.48\textwidth}{!}
	{
        \begin{tabular}{l|c|cccccc|c}
            \toprule
            \multirow{2}{*}{Method} & \multirow{2}{*}{Modality} & \multicolumn{6}{c|}{Per-class IoU} & \multirow{2}{*}{mIoU} \\   \cmidrule(lr){3-8} 
             &  & Dri. & P.C. & Walk. & S.L. & Car. & Div. & \\
            \midrule
             {BEVFusion}~\cite{liu2022bevfusion}& \multirow{2}{*}{LC} & {75.0} & {42.6} & {52.6} & {24.4} & {26.6} & {36.0} & \textcolor{black}{42.9} \\           
            + \name &  & \textbf{78.0} & \textbf{45.9} & \textbf{55.5} & \textbf{26.1} & \textbf{35.4} & \textbf{38.9} & \textcolor{black}{\textbf{46.6}} \\ \midrule 

          {BEVDet}~\cite{huang2021bevdet}& \multirow{2}{*}{C} & {72.7} & {35.6} & {44.7} & {21.1} & {34.0} & {32.3} & {40.1} \\    
             + \name &  & \textbf{76.1} & \textbf{39.9} & \textbf{49.0} & \textbf{23.5} & \textbf{41.6} & \textbf{35.6} & \textbf{44.3} \\
            \bottomrule
        \end{tabular}
        }

        \label{tab:nuscene_seg}
\vspace{-3mm}
\end{table}

\section{Experiment}
\label{sec:experiment}
On multi-modal 3D perception benchmarks (Sec.~\ref{dataset_and_evaluation_metric}), we first verify the transferability of the representation learned via \name{} (Sec.~\ref{main_transfer_results} and Sec.~\ref{label_efficient_transfer_results}). Then, we study the component effectiveness of \name{} (Sec.~\ref{ablation_study}).

\subsection{Dataset and Evaluation Metric}
\label{dataset_and_evaluation_metric}

\mypara{\nus{}}~\cite{Caesar2020nuScenesAM} is a large-scale autonomous driving dataset for 3D perception. Each frame in \nus{} contains six cameras with surrounding views and Lidar point clouds. For {3D object detection}, there are up to 1.4 million annotated 3D bounding boxes for 10 classes; detection score~(NDS) and mean average precision (mAP) across 10 foreground classes are used for evaluation. For {BEV map segmentation}, the Intersection-over-Union (IoU) on 6 background classes and the class-averaged mean IoU (mIoU) are used for evaluation.

\mypara{KITTI-3D}~\cite{kitti} is one of the most widely-used benchmarks for 3D object detection, which consists of 7,481 and 7,518 image-Lidar pairs for training and testing, respectively. The training set is commonly divided into \textit{train} split with 3712 samples and \textit{val} split with 3769 samples~\cite{kittisplit}. As KITTI-3D is a small-scale benchmark existed for a long time, the performance of multi-modal perception models on it is close to saturation. 3D IoU (AP${}_{3D}$) and BEV IoU (AP${}_{BEV}$) with the average precision metric are used for evaluation.

\subsection{Main Transfer Results}
\label{main_transfer_results}In this section, we evaluate the representation quality of pre-training by directly transferring to various perception models on two mainstream 3D perception tasks.

We follow the fine-tuning setups of baseline models and fine-tune the whole framework of them in an end-to-end manner and {{without using any extra data}}. Concretely, we fine-tune models on \nus{} and KITTI-3D for 20 and 80 epochs, respectively. {By default, CBGS~\cite{Zhu2019ClassbalancedGA} trick is not used during fine-tuning.} Then, we {compare the performance between models that are \textit{\textbf{without pre-training}} {and} models whose embedding networks are \textit{\textbf{pre-trained via \name}}.

\subsubsection{Transfer to 3D Object Detection Task}

\mypara{Transfer to Multi-modal Perception Models.} For representation transferability evaluation, we select two baseline models, \textit{i.e.}, BEVFusion~\cite{liu2022bevfusion} and VFF~\cite{li2022voxel}, which represent the state-of-the-art performance for multi-modal 3D perception on \nus{}~\cite{nuscenes} and KITTI-3D benchmarks, respectively. In Table~\ref{tab:nuscene_det}, we show that \name{} can effectively boost the performance for the multi-modal model BEVFusion under various input settings, including varied Lidar sweeps and image resolutions. Notably, for the multi-modal version of the BEVFusion model with multi-sweep Lidar and higher image resolution (\#Sweep:9, \#ImgSize:256$\times$704), \textit{\textbf{\name{} brings} \textbf{more than 2\% improvement (+2.2\%) in mAP} \textbf{and 1.4\% improvement in NDS}}. Moreover, in Table~\ref{tab:kitti}, for VFF~\cite{li2022voxel} framework with various 3D detection heads, including SECOND~\cite{yan2018second}, PVRCNN~\cite{shi2020pv}, and VoxelRCNN~\cite{voxelrcnn}, the representation pre-trained via \name{} also improves the BEV and 3D detection performance in nearly all sub-metrics than training from scratch.

\mypara{Transfer to Single-modal Perception Models.} In Table~\ref{tab:nuscene_det_merged}, we show the generality of pre-trained representation via \name{} to be transferred to single-modal, \textit{i.e.}, camera-only and Lidar-only, perception models. For the camera-only perception model BEVDet\footnote{We use an efficient implementation of BEVDet~\cite{huang2021bevdet} in~\cite{liu2022bevfusion}.}~\cite{huang2021bevdet}, \textit{\textbf{\name{}} \textbf{brings more than 2\% NDS improvement}} for the settings of both the default (20) and doubled ($2\times$: 40) training epochs. For the Lidar-only baselines~\cite{bai2022transfusion,Yin2020Centerbased3O}, the transferable representation learned via our \name{} consistently shows its effectiveness. Moreover, compared to a concurrent work (Voxel-MAE~\cite{voxel-mae}) that targets pre-training for the Lidar-only perception models, our \name{} that targets unified pre-training for multi-modal perception models also shows better performance (62.1\% \textit{v.s.} 61.4\% in NDS) for the Lidar-only baseline, CenterPoint~\cite{Yin2020Centerbased3O}.

\subsubsection{Transfer to BEV Map Segmentation Task}

As is shown in Table~\ref{tab:nuscene_seg}, \name{} largely boosts the performance for BEV map segmentation. In specific, \textit{\textbf{our pre-training schema (\name) largely helps improve the baseline setting with about \textbf{4\%} \textbf{mIoU improvement}}} for both the multi-modal and camera-only perception models. Notably, \name{} brings better segmentation quality for all classes.

\renewcommand{\arraystretch}{0.5}
\begin{table}[t]
\small
\caption{\textbf{Label-Efficient regime 3D object detection (top) and BEV map segmentation (down) results for Multi-modal Perception} model on \nus{} \cite{Caesar2020nuScenesAM} $val$. Models are fine-tuned with varied ratios of annotations. Images of size $256\times704$ and multi-sweep (9) Lidar points are used as input.}\vspace{-2mm}
\setlength{\tabcolsep}{8.6pt}
\centering
\resizebox{0.48\textwidth}{!}
	{
\begin{tabular}{l|c|c|c|c|c}
\toprule
\multirow{2}{*}{Method} & \multirow{2}{*}{Metric} &\multicolumn{4}{c}{Sampling Ratio} \\
\cmidrule(lr){3-6} 
   &  & \multicolumn{1}{c}{  1\%} & \multicolumn{1}{c}{  5\%} & \multicolumn{1}{c}{10\%}  & \multicolumn{1}{c}{100\%}\\ \midrule
         \multicolumn{6}{c}{Label-Efficient regime 3D object detection}       
         \\  
\midrule 
{BEVFusion~\cite{liu2022bevfusion}}& \multirow{2}{*}{mAP} 
              & {26.2}  & {46.1}& {54.2}  & \textcolor{black}{60.8} \\
 {+ \name} && \textbf{{30.2}}  & \textbf{{47.6}}& \textbf{55.9}  & \textbf{{63.0}} \\
 \midrule 
{BEVFusion~\cite{liu2022bevfusion}}& \multirow{2}{*}{NDS} 
              & {44.2}  & {55.4}& {60.3}  & \textcolor{black}{64.1} \\
{+ \name} && \textbf{{45.4}}  & \textbf{{57.0}}& \textbf{61.4}  & \textbf{{65.5}} \\ \midrule \midrule
\multicolumn{6}{c}{Label-Efficient regime BEV map segmentation}       
         \\  \midrule
{BEVFusion~\cite{liu2022bevfusion}}& \multirow{2}{*}{mIoU} & {29.7}  & {39.4}& {41.3}  & \textcolor{black}{42.9} \\
\textcolor{black}{+ \name} && \textbf{{31.1}}  & \textbf{{41.6}}& \textbf{{45.1}}  & \textcolor{black}{\textbf{{46.6}}} \\
\bottomrule
\end{tabular}
}

\label{tab:nuscene_label_efficient_det}
\vspace{-3mm}
\end{table}

\subsection{Label-Efficient Transfer Results}
\label{label_efficient_transfer_results}
Efficient transfer (\textit{e.g.}, fewer annotations) on downstream tasks brings about great application value to real-world scenarios. Thus, we evaluate both label-efficient regime 3D detection and BEV segmentation on \nus.
Specifically, \nus{} training split is sampled with different ratios (1\% to 10\%) to generate label-efficient fine-tuning datasets. For fair comparisons, all models are end-to-end fine-tuned for the same iterations as the default fine-tuning setting (100\%).

 In Table~\ref{tab:nuscene_label_efficient_det}, \textit{\textbf{our pre-training schema} \textbf{shows the best performance under all settings of sampling ratios of labeled fine-tuning data}} for both 3D object detection and BEV-map segmentation tasks, demonstrating good label-efficient transferability of the representation pre-trained via \name.

\subsection{Ablation Study}
\label{ablation_study}

\noindent We ablate the components of \name{} in Table~\ref{tab:ablation_study} and provide the following insights about how to learn transferable multi-modal representation for perception via NS-MAE.

\noindent {\textbf{Masking of input modalities is critical for effective representation learning.}} \textit{\textbf{Without masking:}} When rendering the feature maps of color and depth without masking (settings (A1) and (B3)) for representation learning, the transferred performance could be even worse than the baseline setting (Baseline). 
\textit{\textbf{With masking:}} When enabling the multi-modal masking for the inputs, the comparison between settings (A2)(B1)(B2) with the baseline (Baseline) verifies the effectiveness of masked reconstruction of the feature map of color and projected point cloud, \textit{i.e.}, depth.

\mypara{{Rendering with more modalities and more view directions encourages more transferable representation.}} \textit{\textbf{More modalities:}} When multi-modal masking is enabled, the comparisons between each setting of (A2)(B2)(B3) and the baseline (Baseline) verify the effectiveness of each rendering target ($\mathbf{C}$, $\mathbf{D}^{PER}$, and $\mathbf{D}^{BEV}$). \textit{\textbf{More view directions:}} As point clouds lie in the 3D space, rendered feature maps of projected point clouds ($\mathbf{D}$) in diverse views can be simultaneously optimized in a self-supervised manner. When rendering the projected feature maps of point clouds with more views for reconstruction-based optimization (B4), the performance can be further boosted compared to using only a single view direction ((B1) or (B2)). Finally, by incorporating all rendering targets above with multi-modal masking used, the default setting of \name{} (Default) achieves the best performance with 1.5\% higher NDS and 1.0\% higher mAP than the baseline setting due to their mutual benefits.

\begin{table}[t]
\small
\caption{\textbf{Ablation study} on \nus{} \cite{Caesar2020nuScenesAM} $val$ with the multi-modal perception baseline model, BEVFusion~\cite{liu2022bevfusion}. Results of 3D object detection are reported. $\mathbf{C}$, $\mathbf{D}^{PER}$, and $\mathbf{D}^{BEV}$ denote color, perspective-view depth, and BEV depth. Images of size $128\times352$ and single-sweep Lidar point clouds are used. }\vspace{-1mm}
\setlength{\tabcolsep}{6.pt}
	\centering
        \resizebox{0.48\textwidth}{!}
	{
	\begin{tabular}{l|c|ccc|cc}
          \toprule
      \multirow{2}{*}{Setting}  & \multirow{2}{*}{Masking} &\multicolumn{3}{c|}{Rendering Tagets} & \multirow{2}{*}{mAP} & \multirow{2}{*}{NDS}  \\ \cmidrule(lr){3-5} 
     &     &  $\mathbf{C}$ & $\mathbf{D}^{PER}$& $\mathbf{D}^{BEV}$&   &  \\ \midrule
  Baseline &            &    &   &   &  {50.5} &{53.3}  \\ \midrule
  (A1) &       &$\checkmark$&  &  &{50.2}  &{53.9}\\     
  (A2) &      $\checkmark$&$\checkmark$& &  &{50.7} &{54.2}     
         \\   \midrule
  (B1) &     $\checkmark$  &  & $\checkmark$ & & {{50.9}} &{54.2} \\ 
  (B2) &     $\checkmark$  &  &  &$\checkmark$ & 51.0 & 54.3  \\  
   (B3) &      &  &  $\checkmark$&$\checkmark$ & 49.6 & 52.8\\ 
   (B4) &    $\checkmark$  &  &  $\checkmark$&$\checkmark$   & {{51.3}} &{54.4}\\  \midrule
   Default &      $\checkmark$  &  $\checkmark$ &  $\checkmark$&$\checkmark$  &\textbf{{51.5}} &\textbf{{54.7}}     
         \\  
        \bottomrule
	\end{tabular}
        }
	\label{tab:ablation_study}
\end{table}

\begin{figure*}[t!]
	\centering
 \includegraphics[width=\textwidth]{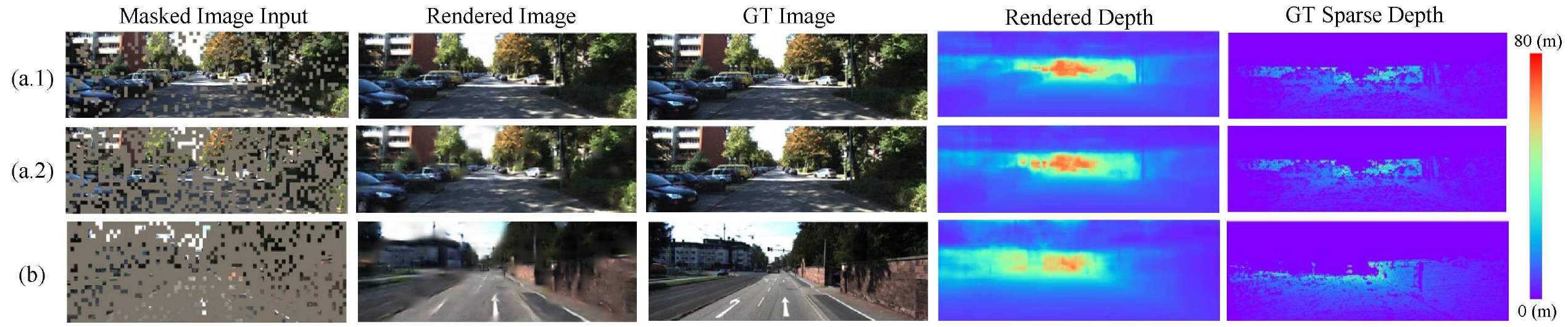}
\vspace{-5mm}
	\caption{\textbf{Qualitative results of perspective-view image and depth}, rendered with front-view camera parameters, on KITTI~\cite{kitti} $val$ set. }
	\label{fig:render_pers}
 \vspace{-5mm}
\end{figure*}

\begin{figure}[t!]
	\centering
 \includegraphics[width=0.4\textwidth]{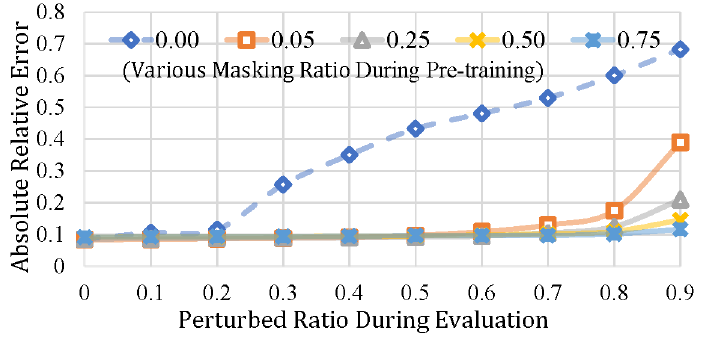} 
 \vspace{-1mm}
	\caption{\textbf{Quantitative comparisons on depth reconstruction quality} (measured by absolute relative error) on KITTI~\cite{kitti} $val$ set with varied perturbed ratios of the input image during evaluation. Here, perturbation is implemented with random masking.}
	\label{fig:perturbed_exp}
  \vspace{-2mm}
\end{figure}

\begin{figure}[t]
	\centering
 \includegraphics[width=0.48\textwidth]{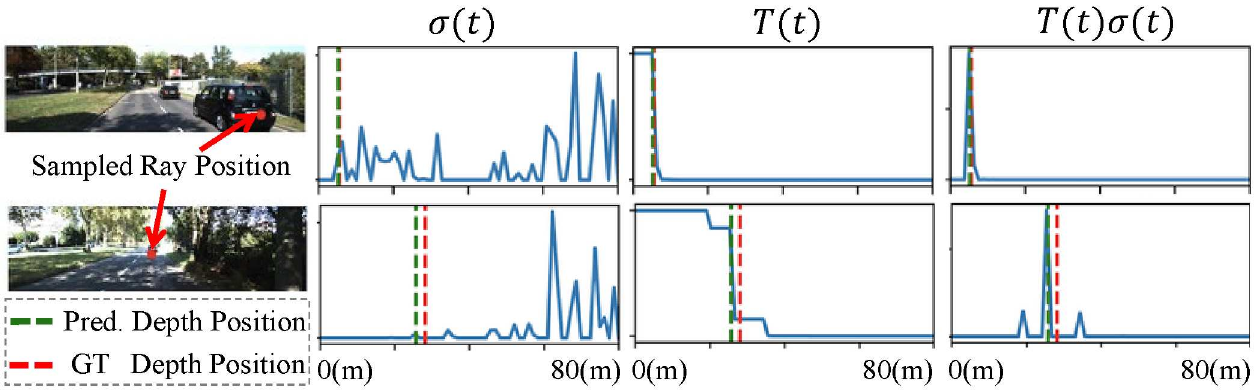} 
 \vspace{-5mm}
	\caption{\textbf{Qualitative visualizations of the radiance terms in rendering equations} (\textit{c.f.} Sec.~\ref{Sec: Radiance Field Rendering}), including volume density $\sigma(t)$, accumulated transmittance $T(t)$, and their product $T(t)\sigma(t)$, of sampled rendering rays on KITTI~\cite{kitti} $val$. Here, the rendering view direction is orthogonal to the perspective-view image plane. }
	\label{fig:render_equation}
\end{figure}

\subsection{Analysis}
In this section, we analyze the emerging properties of \name{} for multi-modal perception pre-training.

\mypara{\name{} Induces Robust Multi-modal Reconstruction.}
In Fig.~\ref{fig:render_pers}, given the input images with diverse masking ratios, we show the rendering results of perspective-view images and projected point cloud feature maps, \textit{i.e.}, depth maps. Case (a.1)\&(a.2) of Fig.~\ref{fig:render_pers} shows that the multi-modal embedding network pre-trained with \name{} can render visually-high-quality color and depth of the scene. In Case (b) of Fig.~\ref{fig:render_pers}, even if the input image is masked out with an extremely high masking ratio (80\%), the rendered image and depth can still reconstruct coarse-level color and geometry well. To quantitatively study the robustness of the multi-modal representation pre-trained via \name, we provide the depth reconstruction results of models trained and evaluated with various masking ratios on KITTI~\cite{kitti}. As is shown in Fig.~\ref{fig:perturbed_exp}, the models trained with masked modeling exhibit better robustness to perturbations thanks to the learned modality completion and generalization ability. 

\mypara{\name{} Is Physics-informed Unknown-region Filter.}
In Fig.~\ref{fig:render_equation}, we visualize the radiance terms in rendering equations (Eq.~(\ref{equ:render_color_d}) and Eq.~(\ref{equ:depth_render_d}) in Sec.~\ref{Sec: Radiance Field Rendering}) of sampled rendering rays to understand the property of the rendering equations better. During reconstruction-based optimization (Sec.~\ref{Sec: Multi-modal Reconstruction}), the accumulated transmittance $T(t)$ term in the prior rendering functions serves as a \textit{band-pass filter} along the ray direction, \textit{e.g.}, the direction from the origin (0m) to the farthest (80m) of Fig.~\ref{fig:render_equation}. Specifically, $T(t)$ helps {selectively filter} out the \textit{\textbf{unknown}} region (from the position of an object to the farthest, \textit{i.e.}, \textit{approximate infinity}), and only {optimize} \textit{\textbf{non-occupied}} regions (from the origin of the ray to the object) and \textit{\textbf{occupied}} regions (where objects exist), respectively. Besides, the sigma field $\sigma(t)$ term is learned to predict zero for \textit{\textbf{non-occupied}} regions between the origin of the ray and the object, deriving a focused impulse in the product of accumulated transmittance and sigma field $T(t)\sigma(t)$.

\mypara{\name{} Generates Dense BEV Occupancy.}
Given an \textit{\textbf{extremely sparse}} masked Lidar input, the pre-training schema of \name{} includes the sub-task of reconstructing the projected BEV point cloud feature maps, \textit{i.e.}, BEV depth, which indicates the occupancy of objects on the BEV plane. In Fig.~\ref{fig:render_bev}, we show that \name{} pre-trained with only image and Lidar pairs can help the multi-modal perception model learn to complete \textit{\textbf{denser}} BEV occupancy maps which decouple objects in the scene along the height axis of the world coordination. Such a good zero-shot BEV occupancy completion ability can further elevate the performance on BEV perception downstream tasks, especially for BEV map segmentation (as is verified in the experiments of Sec.~\ref{main_transfer_results}).

\begin{figure}[t]
	\centering
 \includegraphics[width=0.48\textwidth]{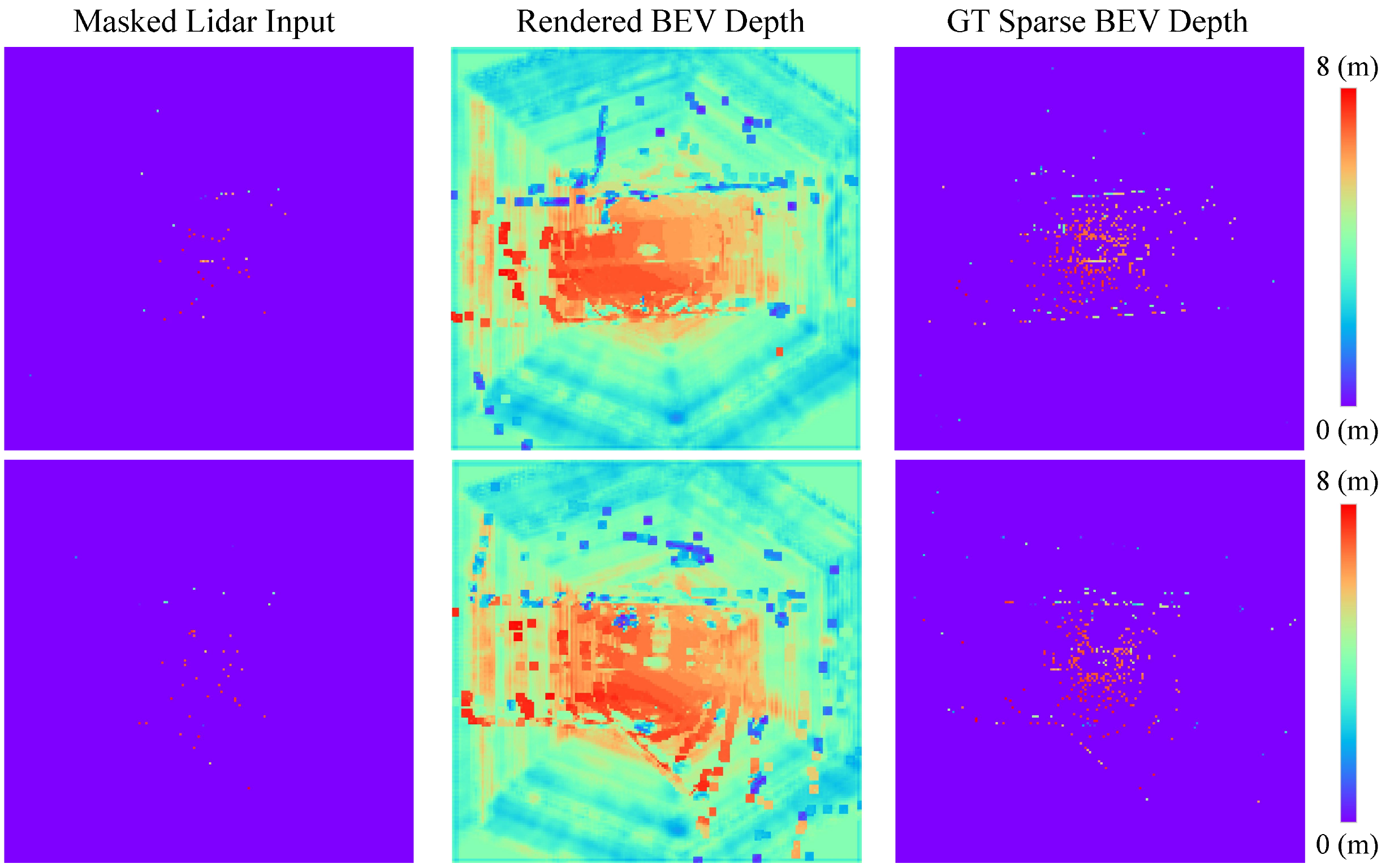} 
\vspace{-5mm}
	\caption{\textbf{Qualitative results of bird's-eye-view (BEV) depth}, \textbf{\textit{i.e.}, BEV occupancy}, rendered from a viewpoint that is orthogonal to the road plane, on \nus ~\cite{nuscenes} $val$. \textbf{Zoom in to view better.}}
	\label{fig:render_bev}
\end{figure}

To sum up, \name{} naturally inherits useful properties from MAE and NeRF, thus learning transferable representation for multi-modal perception models effectively.

\section{CONCLUSIONS AND FUTURE WORK}
\label{sec:Conclusion}
We have proposed a unified self-supervised pre-training paradigm (\name) for multi-modal perception models. Specifically, \name{} conducts masked multi-modal reconstruction in NeRF to enable transferable multi-modal representation learning in a unified and neat fashion.
Extensive experiments demonstrate the encouraging transferability and generalization of the learned representation via \name{} for both multi-modal and single-model perception models. 

\mypara{Future Work.} Due to computation and time limitations, we currently do not explore \name{} with larger models and data. Besides, we consider it meaningful to study the generality of \name{} to perception models with more modalities, \textit{e.g.}, Radar~\cite{nabati2021centerfusion,li2022exploiting,yang2020radarnet}, or joint multi-modal encoders~\cite{yang2022deepinteraction}.

\section{Acknowledgements}
The author would like to thank Shusheng, Zhaoyang, Baixin, Huaxin, and Yiheng for their kind help with proofreading. Also, the author would like to thank Yangyi, Chenyu, Jinguo, and Tianyu for their valuable viewpoints and help during the development of this project. Moreover, the author is sincerely grateful to the AI Data Center (AIDC) in Shanghai Lingang for providing access to GPU resources, and would like to express their heartfelt acknowledgment.

\renewcommand{\thetable}{{\Alph{table}}}
\renewcommand{\thefigure}{{\Alph{figure}}}
\renewcommand{\thesection}{{\Alph{section}}}
\setcounter{section}{0}  
\setcounter{figure}{0}  
\setcounter{table}{0}  

\newlength\savewidth\newcommand\shline{\noalign{\global\savewidth\arrayrulewidth
  \global\arrayrulewidth 1pt}\hline\noalign{\global\arrayrulewidth\savewidth}}
\newcommand{\tablestyle}[2]{\setlength{\tabcolsep}{#1}\renewcommand{\arraystretch}{#2}\centering\footnotesize}

\section{More Qualitative Comparisons}

As is shown in Fig.~\ref{fig:bev_map_seg_result}, we compare the BEV map segmentation results predicted by the multi-modal perception model~\cite{liu2022bevfusion} when \textit{trained from scratch} and the model \textit{pre-trained via the proposed NS-MAE} self-supervised representation learning schema, which qualitatively demonstrates that the transferred representation learned via NS-MAE can help
improve the segmentation quality for 3D perception. 

\begin{figure}[t]
	\centering
 \includegraphics[width=0.48\textwidth]{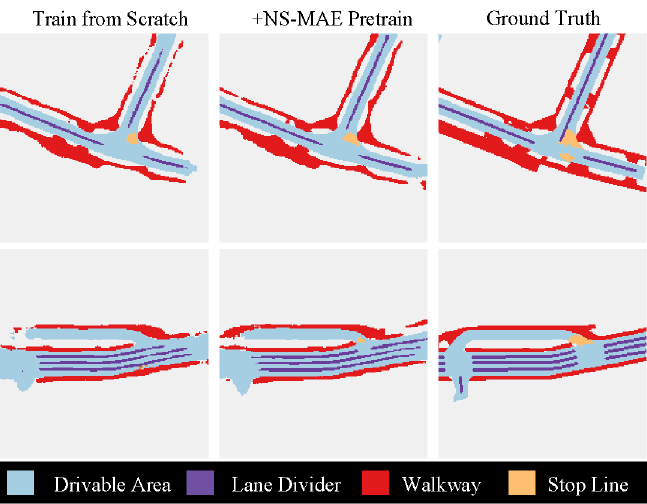} 
 \vspace{-4mm}
	\caption{\textbf{Qualitative comparisons of BEV map segmentation results} on \nus ~\cite{nuscenes} $val$ with multi-modal perception model BEVFusion~\cite{liu2022bevfusion}. Specifically, we compare the qualitative results of the model which is {trained from scratch} (Train from Scratch) and the results of the model which is {pre-trained via NS-MAE} (+NS-MAE Pretrain). The ground-truth BEV map segmentation result (Ground Truth) is provided for reference.}
	\label{fig:bev_map_seg_result}
\end{figure}

\section{More Details for Baseline Models}

\subsection{Multi-modal Perception Model}

\mypara{BEVFusion~\cite{liu2022bevfusion}} is an advanced fusion-based/multi-modal perception model for both 3D object detection and BEV map segmentation. (1) For the embedding network part, BEVFusion uses Swin-T~\cite{liu2021swin} with FPN~\cite{lin2017feature} as the Camera Encoder, VoxelNet~\cite{yan2018second} as Lidar Encoder, and LSS~\cite{philion2020lift} as Cam2World view-transformation module. (2) In the original implementation of BEVFusion, the Lidar point cloud is voxelized with different resolutions for diverse downstream tasks, \textit{i.e.},  0.075(m) (for 3D object detection) and 0.1(m) (for BEV map segmentation). Differently, to unify and align the multi-modal pre-training and the multi-modal transferring, we voxelize the Lidar point cloud with the same resolution, \textit{i.e.}, 0.075(m), for both detection and segmentation tasks. (3) Due to computation and time limitations, although CBGS~\cite{Zhu2019ClassbalancedGA} data augmentation is an effective trick to improve the performance for downstream tasks, we do not use it for all the experiments in the main paper. (4) For the implementation of the rendering network, we use separate \textit{conv} layers for the bird's eye view and perspective views. (5) We refer to the detailed implementations in the official repo of BEVFusion, which is under the license of \href{https://github.com/mit-han-lab/bevfusion/blob/main/LICENSE}{\texttt{Apache 2.0}}, for the pre-training, fine-tuning, and evaluation.

\mypara{VFF~\cite{li2022voxel}}, \textit{namely} Voxel Field Fusion, is a perception model for cross-modality 3D object detection. (1) For the embedding network part, VFF uses ResNet-50~\cite{he2016deep} as the Camera Encoder and the proposed voxel fusion mechanism for Cam2World transition and fusion. For the Lidar Encoder, VFF is instantiated with various voxel-based Lidar backbones, \textit{e.g.},
PV-RCNN~\cite{pvrcnn} and VoxelRCNN~\cite{voxelrcnn}. (2) In the VFF-SECOND configuration that we set up for transferability evaluation of the representation pre-trained via NS-MAE in the main paper, we refer to the baseline model setting of their paper for re-implementation. Specifically, we remove the voxel-field-fusion mechanism from the default setting of VFF and use the 3D detection head of SECOND~\cite{yan2018second}. (3) For the rendering network, we implement it with separate \textit{conv} layers for the bird's eye view and perspective views. Specifically, we empirically leverage FPN-like~\cite{lin2017feature} convolution layers for the perspective-view rendering network and simple conv$3\times 3$ layer for the bird's-eye-view rendering network. (4) We refer to the detailed implementations in the official repo of VFF, which is under the license of \href{https://github.com/dvlab-research/VFF/blob/main/LICENSE}{\texttt{Apache 2.0}}, for the pre-training, fine-tuning, and evaluation.

\subsection{Camera-only Perception Model}

\mypara{BEVDet~\cite{huang2021bevdet}} is a sophisticated camera-only perception model for 3D object detection. We leverage an efficient implementation of BEVDet in~\cite{liu2022bevfusion} to align the setting between the multi-modal representation learning during pre-training and the fine-tuning stage. (1) For the embedding network part, BEVDet uses Swin-T~\cite{liu2021swin} with FPN~\cite{lin2017feature} as the Camera Encoder and leverages LSS~\cite{philion2020lift} as the Cam2World view-transformation module. (2) During the pre-training of BEVDet, we additionally leverage the CBGS~\cite{Zhu2019ClassbalancedGA} data augmentation, which elongates the training iterations to 4.5 times than the default setting without it, for the convergence of the model. (3) We refer to an improved implementation of BEVDet in the repo of BEVFusion, which is under the license of \href{https://github.com/mit-han-lab/bevfusion/blob/main/LICENSE}{\texttt{Apache 2.0}}, for the pre-training, fine-tuning, and evaluation.

\subsection{Lidar-only Perception Model}
\mypara{CenterPoint~\cite{Yin2020Centerbased3O}} is a widely-used Lidar-only perception model for 3D object detection. (1) For the embedding network, CenterPoint uses a standard Lidar-based backbone network, \textit{i.e.}, VoxelNet~\cite{Zhou2018VoxelNetEL} to build a representation of the input point cloud. (2) We refer to the implementation in the open-sourced MMDetection3D \cite{mmdet3d} library, which is under the license of \href{https://github.com/open-mmlab/mmdetection3d/blob/master/LICENSE}{\texttt{Apache 2.0}}, and the official repo of CenterPoint, which is under the license of \href{https://github.com/tianweiy/CenterPoint/blob/master/LICENSE}{\texttt{MIT}}, for the pre-training, fine-tuning, and evaluation.

\section{More Details for Evaluation Protocal}
\mypara{3D object detection on \nus}: We leverage the official overall metrics: mean Average Precision (mAP) \cite{Everingham2009ThePV} and {\nus{} Detection Score} (NDS), and report these two metrics in the main paper for performance comparisons. Here, we detail the calculation of them as follows. In specific,  mAP is computed by averaging over distance thresholds of 0.5(m), 1(m), 2(m), and 4(m) across ten sub-classes: car, truck, bus, trailer, construction vehicle, pedestrian, motorcycle, bicycle, barrier, and traffic cone. Moreover, \nus{} Detection Score (NDS) is a consolidated metric of mAP and all other indicators (e.g., translation, scale, orientation, velocity, and attribute) to comprehensively judge the 3D detection quality on \nus{} dataset.

\mypara{BEV map segmentation on \nus}: Following the evaluation protocol on the BEV map segmentation task~\cite{philion2020lift,li2022bevformer,liu2022bevfusion,liang2022bevfusion}, the region for evaluation is limited in the [-50m, 50m]$\times$[-50m, 50m] region around the ego car.

\section{Liscenses of Datasets}
\mypara{nuScenes~\cite{nuscenes}} is a large-scale 3D perception dataset released under the \href{https://www.nuscenes.org/terms-of-use}{\texttt{CC BY-NC-SA 4.0}} license.

\mypara{KITTI-3D~\cite{kitti}} is a small-scale 3D perception dataset released under the the \href{https://www.cvlibs.net/datasets/kitti/}{\texttt{CC BY-NC-SA 3.0}} license.

\bibliographystyle{ieee_fullname}
\bibliography{main}


\end{document}